\documentclass[preprint,5p,times]{elsarticle}
\usepackage[
colorlinks=true,
linkcolor=blue,
urlcolor=blue,
citecolor=blue,
pdftex,
]{hyperref}
\usepackage{amsmath,graphicx,epstopdf,tikz}
\usepackage{enumitem}
\usepackage{dsfont}
\usepackage[numbers]{natbib} 

\usepackage{subcaption}

\usepackage{multirow}
\usepackage{array}
\usepackage{color}
\usepackage{colortbl}
\usepackage{tabularx}
\usepackage{algpseudocode}
\usepackage{algorithm}
\usepackage{enumerate}

\usepackage{booktabs}
\usepackage{rotating}
\graphicspath{{figures/}}
\usepackage{threeparttable}

\usepackage{todonotes}

\newcolumntype{"}{@{\hskip5pt\vrule width 1pt\hskip\tabcolsep}}
\makeatother

\journal{Pervasive and Mobile Computing}

\begin{document}

\begin{frontmatter}



\title{Unsupervised Understanding of Location and Illumination Changes in Egocentric Videos}


\author{
    \begin{tabular*}{0.9\textwidth}{c
@{\extracolsep{\fill}} c @{\extracolsep{\fill}} c} Alejandro Betancourt$^{1,2}$& Natalia D\'iaz-Rodr\'iguez$^{3}$ & Emilia Barakova$^{2}$ \\ 
\texttt{ \small a.betancourt@tue.nl} & \texttt{ \small ndiaz@decsai.ugr.es} & \texttt{\small e.i.barakova@tue.nl} \\
\\
Lucio Marcenaro$^{1}$& Matthias Rauterberg$^{2}$ & Carlo Regazzoni$^{1}$ \\
\texttt{\small lucio.marcenaro@unige.it} & \texttt{\small g.w.m.Rauterberg@tue.nl} & \texttt{\small carlo@dibe.unige.it} \\ \\ \\
\end{tabular*}}

\address{
\begin{tabular*}{0.9\textwidth}{c
@{\extracolsep{\fill}} c @{\extracolsep{\fill}} c}
    $^1$ \small Department of Engineering (DITEN). & \small  $^2$ Department of Industrial Design. & \small   $^3$ Computer Science Department. \\ 
\small University of Genova &  \small Eindhoven University of \small Technology. &  \small University of California Santa Cruz  \\ 
\small Genova, Italy &  \small Eindhoven, Netherlands. &  \small California, USA. \\ 
\end{tabular*}}

\begin{abstract}
Wearable cameras stand out as one of the most promising devices for the upcoming years, and as a consequence, the demand of computer algorithms to automatically understand the videos recorded with them is increasing quickly. An automatic understanding of these videos is not an easy task, and its mobile nature implies important challenges to be faced, such as the changing light conditions and the unrestricted locations recorded. This paper proposes an unsupervised strategy based on global features and manifold learning to endow wearable cameras with contextual information regarding the light conditions and the location captured. Results show that non-linear manifold methods can capture contextual patterns from global features without compromising large computational resources. The proposed strategy is used, as an application case, as a switching mechanism to improve the hand-detection problem in egocentric videos.
\end{abstract}

\begin{keyword}
   Machine Learning \sep Unsupervised Learning \sep Egocentric Videos \sep First Person Vision, Wearable Camera
\end{keyword}

\end{frontmatter}

\section{Introduction} \label{sec:intro}

The emergence of wearable video devices such as action cameras, smart glasses and low-temporal life-logging cameras has detonated a recent trend in computer science known as First Person Vision (FPV) or Egovision. The 90's idea of a wearable device with autonomous processing capabilities is nowadays possible and is considered one of the most relevant technological trends of the recent years \cite{Betancourt2014}.  The ubiquitous and personal nature of these devices opens the door to critical applications such as Activity Recognition \cite{Nguyen2016, Zhan2015}, User-Machine Interaction \cite{Baraldi2015}, Ambient Assisting Living \cite{Fathi2011, Pirsiavash2012, NataliaDiaz2014} Augmented Memory \cite{Farringdon2000, Harvey2016} and Blind Navigation \cite{Balakrishnan2007}, among others. 

One of the key features of wearable cameras is their capability to move across different locations and record exactly what the user is looking at. This is an unrestricted video perspective that requires existent methods to perform good in the unknown number of locations and the changing light conditions implied by this video perspective. A common way to deal with this problem is to predefine a particular application or location and bound the algorithms based on this. This is the case of gesture recognition for virtual museums proposed in \cite{Baraldi2015} or the activity recognition methods based on the kitchen dataset \cite{Fathi2011,Fathi2011a}. Another way to alleviate the large number of recorded locations is by using exhaustive video labeling of the recorded locations and objects as is done in \cite{Pirsiavash2012} to detect daily activities. The authors in \cite{Li2013b} use global histograms of color to reduce the effect of light changes in a color-based hand-segmenter. 

The approach of \cite{Li2013b} shows that contextual information, such as light conditions, are valuable sources of information that can be used to improve the performance and applicability of current FPV methods. This idea is also applicable to other FPV related functionalities such as activity recognition, on which a device that can understand user's location can easily reduce the number of possible activities and take more accurate decisions. Pervasive computing refers to the devices that can modify their behavior based on contextual variables as context-aware devices \cite{Lara2013}, and its benefits are widely explored for example in assisted living \cite{Riboni2011} and anomaly detection \cite{Zhu2013}.

\begin{figure}[h]
    \centering
    \includegraphics[width=0.7\linewidth]{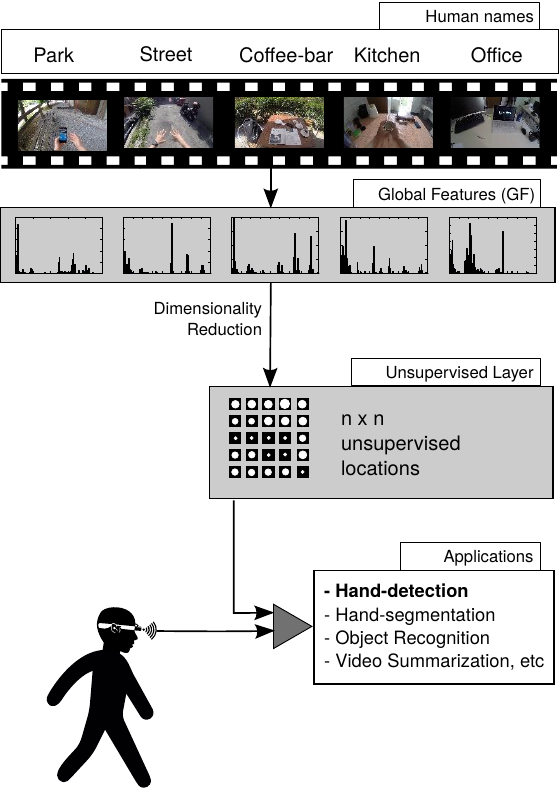}
    \caption{Unsupervised strategy to extract contextual information about
    light and location using global features.}
    \label{diagram}
\end{figure}

This paper is motivated by the potential impact of contextual information, such as light conditions and location, on different FPV methods. The strategy presented, is a first step towards our envision of a device that can understand the environment of the user and modify its behavior accordingly. The proposed approach understands the contextual information on which the user is involved as a set of different characteristics that can point to previously recorded conditions, and not as a scene classification problem based on manual labels assigned to particular locations (e.g., kitchen, office, street). In this way, this study devises an unsupervised procedure for wearable cameras to switch between different models or search spaces according to the light conditions or location on which the user is involved. Figure \ref{diagram} summarizes our approach.

From Figure \ref{diagram} it is clear that the transition from the global features to the unsupervised layer can be seen as a dimensional reduction from the global feature space (high dimensional space) to a simplified low dimensional space (intrinsic dimension). The latter provides an unsupervised location map to be used later to switch between different behaviours at different hierarchical levels. These dimensional reductions are known as manifold methods, and their capabilities to capture complex patterns are defined by their algorithmic and/or theoretic formulation \cite{Friedman1997}.  

Regarding the global features to be used, relevant information can be obtained from recent advances in FPV \cite{Betancourt2014} and scene recognition \cite{Zhou2014, Oliva2001}.  Given the restricted computational resources of wearable devices, we use computationally efficient features such as color histograms and GIST descriptors. However, the proposed approach can be extended with more complex data such as deep features \cite{Zhu2013}. In that case three important issues must be considered: i) the computational cost will restrict the applicability in wearable devices; ii) it will require large amounts of training videos and manual labels; iii) the use of existent ``pre-trained'' neural architectures compromises the unsupervised nature of our approach.

The novelties of this paper are three folded: i) It evaluates the capability of different linear and non-linear manifold methods, namely Principal Component Analysis (PCA), Isometric Mapping (Isomaps), Self Organizing Maps (SOM) and Growing Neural Gas (GNG), to capture light/location patterns from different global features without using manual labels. ii) It analyzes, following a feature selection procedure, the most discriminative components of the selected global features, iii) As an application case, the proposed unsupervised strategy is used to improve the \emph{hand-detection} problem in FPV. The hand-detection problem is used as an example, because of its impact on context-aware devices in hand-based methods, and because it allows us to illustrate the role of the unsupervised layer and its contribution to the final hand-detection performance. The use of the same strategy at higher inference levels such as hand-segmentation or hand-tracking is left as future research.

The remainder of this paper is organized as follows: Section \ref{sec:stateoftheart} summarizes some recent strategies to understand automatically contextual information. Later, Section \ref{sec:method} introduces our methodological approach, summing up the selected features, different manifold methods and some common unsupervised evaluation procedures. In Section \ref{sec:results} the manifold methods are trained, and their capability to capture light/location patterns is evaluated in a post-learning strategy using the manual labels of two public FPV datasets. Section \ref{sec:application} illustrates the use of the best performing manifold method to improve the hand-detection rate in FPV.  Finally, Section \ref{sec:conclusions} concludes and provides some future research lines.

\section{State of the Art}\label{sec:stateoftheart}

In recent years, FPV video analysis is attracting the interest of the researchers, due to the increasing availability of wearable devices that can record what the user is looking at, and promising applications are emerging. Existing literature and commercial approaches highlight a broad range of possibilities, but also points to several challenges to be faced such as uncontrolled locations, illumination changes, camera motion, object occlusions, processing capabilities, among others \cite{Betancourt2014}. This paper addresses the issue of illumination changes as well as unrestricted locations recorded by the camera. The general idea is to develop an unsupervised layer that, based on global features and using low computational resources, understands contextual information regarding the light conditions and the locations recorded by the camera.

The advantages of a device that can understand the environment are evident \cite{Starner1998, Zhu2011}. Recent advances in pervasive computing and wearable devices frequently point at the location of the user as a valuable information source to design context-aware systems \cite{Riboni2011, Lara2013, Wang2011}. An intuitive way to find the location is to use Global Positioning Systems (GPS). However, this approach is commonly restricted by the battery life as well as by poor indoor signal \cite{Hori2003}. 

To alleviate these restrictions, wearable cameras emerge as a possible solution: infer the context using the recorded frames. As an example, in \cite{Templeman2014} local and global features are combined to identify private locations and avoid recording them. In fact, the idea pursued by the authors is in line with the seminal works on scene recognition proposed by Oliva and Torralba, on which scenes captured by static cameras are represented as low dimensional vectors known as GIST descriptors \cite{Oliva2005, Oliva2001} and classified in a supervised way. Recent advances in scene recognition made by the same authors by exploiting the hidden layers of deep networks (deep features) are promising \cite{Zhou2014}. However, their applicability on wearable devices is still restricted by the required computational resources and by the unavailability of large datasets recorded with wearable cameras. 

Similar applications but following an unsupervised strategy are common in robotics, on which manifold algorithms like SOM or Neural Gas, are frequently used in autonomous navigation systems \cite{Puliti2003, Barakova2005, Barakova2005a}. Regarding FPV, the authors in \cite{Li2013b} propose a multi-model recommendation system for hand-segmentation in egocentric videos that modify its internal behaviour based on the recorded light conditions. In their paper, the authors design a performance matrix containing one row per training frame and one column per model. The matrix values are the segmentation scores and are used to decide the most suitable model for each frame in the testing dataset. 

The proposed method is motivated by the switching mechanisms developed by \cite{Li2013b}; however, it is independent on the segmentation dataset and can extract information about the light conditions as well as the recorded location. Regarding the scene-recognition literature, our approach is fully unsupervised and is based on computationally efficient global features which make feasible to use it on wearable cameras.

\section{Unsupervised method}\label{sec:method}

\begin{figure}
    \centering
    \includegraphics[width=1\linewidth]{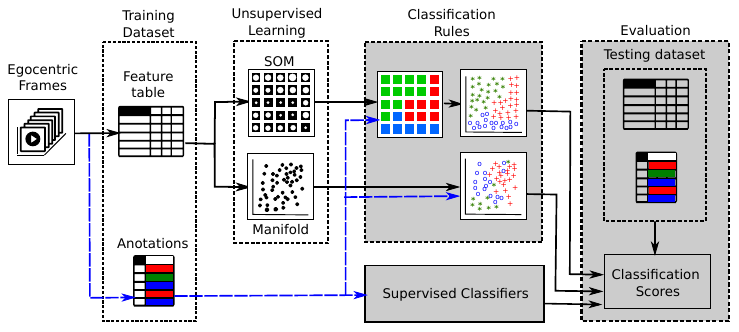}
    \caption{General workflow of our unsupervised evaluation. White blocks correspond to the unsupervised-learning. The manually labeled data is used only by shaded blocks, which correspond to the post-learning evaluation.}\label{fig:diagram}
\end{figure}

As explained in previous sections one of our goals is to quantify the capability of different unsupervised manifold methods to capture the illumination and location changes in egocentric videos. Our approach follows the experimental findings of previous works, on which global features such as color histograms and GIST are used to describe the general characteristics of the scene \cite{Li2013b, Oliva2001}.  Figure \ref{fig:diagram} summarizes our approach. Feature extraction and unsupervised training modules can be found in the left part of the picture, while the right part shows the post-learning evaluation. Manual labels are used in the shaded blocks of the diagram only. The remainder of this section introduces the datasets, motivates the global features and manifold methods, and concludes explaining the hyperparameter selection and the post-learning analysis.

\subsection{Datasets}

The comparison of the manifold methods uses two popular FPV datasets, namely EDSH and UNIGE-HANDS. The main criteria for the dataset selection are the number of locations, the existence labels, and the illumination changes contained. To the best of our knowledge, these datasets are commonly used to compare hand-segmentation algorithms in FPV due to their challenging light conditions intentionally included in the dataset design phase.

\textbf{EDSH:} {Dataset proposed by \cite{Li2013b} to train a pixel-by-pixel Hand-Segmenter in FPV. The dataset contains $8$ different locations with changing light conditions recorded from a head-mounted camera with a resolution of $720p$ at a speed of $30$ \emph{fps}. The labels about location and light conditions are manually created. For the experimental results, EDSH1 video is used for training and EDSH2 video for testing. In total $2806$ frames are used for training and $1067$ for testing. Figure \ref{fig:EDSHcomposition} shows the EDSH training and testing dataset composition according to the labels to be used in the Section \ref{sec:classification}.

\begin{figure}[h!]
    \centering
    \includegraphics[width=0.6\linewidth]{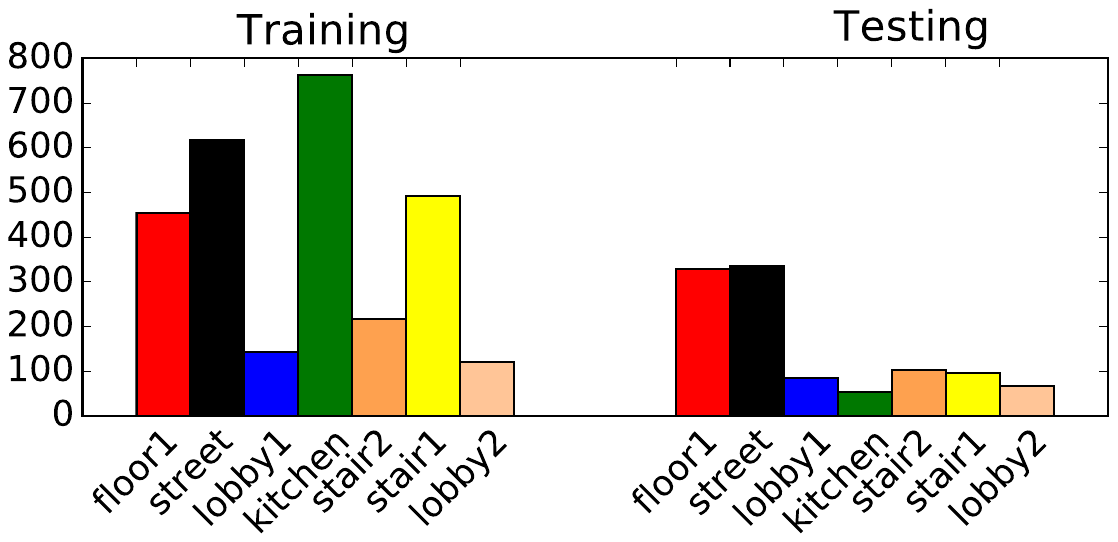}
    \caption{EDSH Training and Testing dataset composition}\label{fig:EDSHcomposition}
\end{figure} 
}
\textbf{UNIGE-HANDS:}{Dataset proposed by \cite{Betancourt2015a} as baseline for the hand-detection problem in FPV. The dataset is recorded in $5$ different locations (1.  Office, 2. Coffee Bar, 3. Kitchen, 4. Bench, 5.  Street), and is recorded with a resolution of $1280 \times 720$ \emph{pixels} and $50$ \emph{fps}. The dataset provides the locations of the videos.  Labels about indoor/outdoor information were manually created. In Section \ref{sec:results} the original training/testing split is used. In total $4436$ frames are used for training and $1406$ for testing. Figure \ref{fig:UNIGEcomposition} shows the EDSH training and testing dataset composition according to the labels to be used in the Section \ref{sec:classification}.

\begin{figure}[h!]
    \centering
    \includegraphics[width=0.6\linewidth]{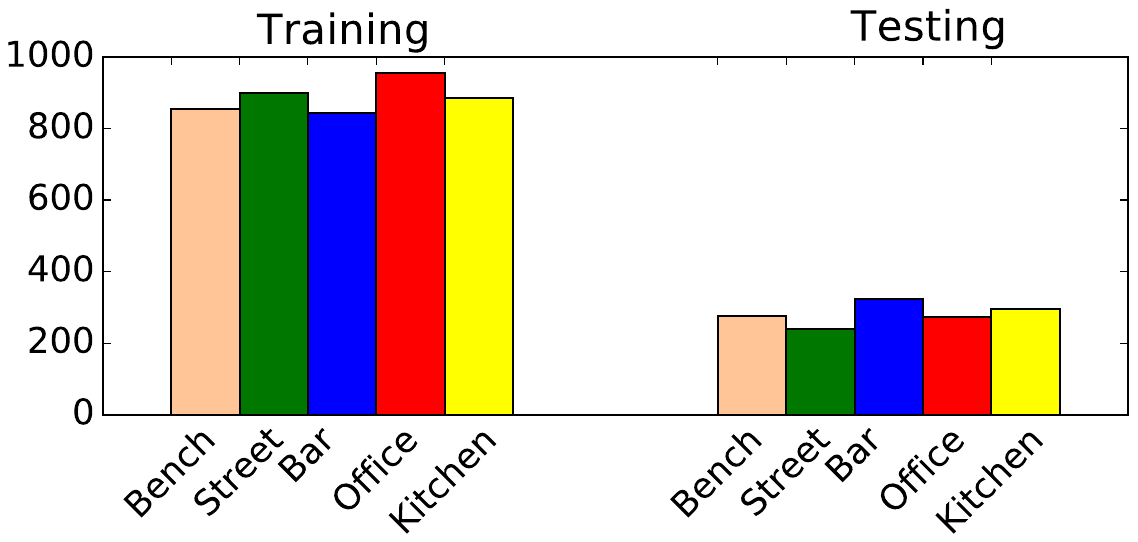}
    \caption{UNIGE Training and Testing dataset composition}\label{fig:UNIGEcomposition}
\end{figure} 
}

\subsection{Feature selection}

To represent the scene context we use color histograms and GIST descriptors. These features are widely accepted and used in the FPV literature, and their computational cost makes them suitable for wearable devices with highly restricted processing capabilities and battery life. As explained before, more complex features such as deep features can be used under the same framework, but different issues must be faced to reach a real applicability. We point deep features as a promising future work.

Due to its straightforward computation and intuitive interpretation, color histograms are probably the most used features in image classification \cite{Kakumanu2007}. The variety of color spaces such as RGB, HSV, YCbCr or LAB makes it possible to exploit color patterns while alleviating potential illumination issues. In particular, HSV is based on the way humans perceive colors while LAB and YCbCr use one of the components for lightness and the remaining ones for the color intensity. In egocentric vision, \cite{Morerio2013, Betancourt2016} use a mixture of color histograms and visual flow for \emph{hand-segmentation}, while \cite{Baraldi2015} combined HSV features, a Random Forest classifier and super-pixels for gesture recognition. Recently, Li and Kitani \cite{Li2013b} analyzed the discriminative power of different color histograms with a Random Forest regressor. Existent FPV literature commonly points to HSV as the best color space to face the changing light conditions in egocentric videos \cite{Morerio2013, Li2013b}. For the experimental results, we use color histograms of RGB, HSV, YCbCr and LAB.

Additionally, we use GIST \cite{Murphy2006} as a global scale descriptor. It captures texture information, orientation and the coarse spatial layout of the image. GIST can be combined with other local descriptors to detect accurately objects in the scene, and was initially combined with a simple one-level classification tree, as well as with a na\"{\i}ve Bayesian classifier. GIST descriptor has been successfully applied on large scale image retrieval and object recognition \cite{Oliva2001}.

Finally, the experimental results analyze the discriminative power, regarding light and location, of the proposed global features under a feature selection procedure. The idea behind this experiment is to fuse the more discriminative components of each global feature to increase the contextual information available in the high-dimensional space, and as consequence improve the patterns captured by the manifold method. For this purpose, all the proposed global features are merged and used with a Random Forest to solve the classification problems explained in Section \ref{sec:results}. The feature importance of the Random Forest is used to build a combined feature with the most discriminative components.

\subsection{Manifold learning}

Manifold methods are mathematic or algorithmic procedures designed to move from a high dimensional space to a low dimensional one while preserving the most valuable information \cite{Friedman1997}. Manifold methods are widely used and its applicability is fully validated in several field such as robotics \cite{Puliti2003, Barakova2005, Barakova2005a}, crowd analysis \cite{Chiappino2014a, morerio2012people} and speech recognition \cite{Arous2010a}, among others.

In general, the capability of manifold methods to deal with complex data is defined by their mathematic formulations and assumptions. Manifold methods are usually grouped according to two factors: i) If the dimensional mapping uses manual labels, then the method is supervised; otherwise, it is unsupervised. As an example, Linear Discriminant Analysis (LDA) and Principal Component Analysis (PCA) are supervised and unsupervised, respectively. ii) If the intrinsic dimensions are linear combinations of the original space then it is linear; otherwise, it is non-linear.  As an example, PCA is linear, and SOM is non-linear. Due to the final objective of this paper, the remaining part does not consider the supervised approaches such as LDA.

To find a well performed dimensional mapping, we use as baseline the Principal Component Analysis (PCA) algorithm, which is the most common linear manifold algorithm but usually fails to capture patterns in complex datasets. To capture complex patterns we use three non-linear manifold methods, namely Isomaps, SOM and GNG. These non-linear algorithms were chosen based on the advantages reported in previous studies \cite{Tenenbaum00, Kohonen1990, Florez2002}, and their capability to be applied to new observations not included in the training data. In our exploratory analysis t-SNE was also used; however, its original formulation cannot be applied to data outside of the training dataset. Regarding SOM and GNG, this study is based on the original formulation to keep simple the interpretation and analysis of the results.

\textbf{Principal Components Analysis:} it is a linear technique to reduce data dimensionality by transforming the original data into a new set of variables that summarize the original data \cite{Tenenbaum00}.  The new variables are the principal components (PCs), and are uncorrelated and ordered such that the $k-th$ PC has the $k-th$ largest variance among all PCs, and the $k-th$ PC is orthogonal to the first $k-1$ PCs. The first few PCs capture the main variations in the dataset, while the last PCs capture the residual ``noise'' in data.

\textbf{Isomaps:} a non-linear dimensionality reduction algorithm proposed in \cite{Tenenbaum00} that learns the underlying global geometry of a dataset using local distances between the observations. In comparison with classical linear techniques, Isomaps can handle complex non-linear patterns such as those in human handwriting or face recognition in images. Isomaps combine the major algorithmic features of PCA and the multidimensional-scaling computational efficiency, global optimality, and asymptotic convergence, which makes feasible its use in wearable cameras. The hyperparameter of Isomaps is the number of neighbors \cite{Jing2011}.

\textbf{Self Organizing Maps (SOM):} it is one of the most popular unsupervised neural networks. It was originally proposed to visualize large dimensional datasets \cite{SOM} and easily find relevant information \cite{SOM01} on them. In summary, the SOM is a two layer neural network that learns a non-linear projection of a high dimensional space (input layer) to a regular discrete low-dimensional grid of neural units (output layer). The discrete nature of the output layer facilitates the visualization of learned patterns and makes easy to find topological relations in the data.

The training phase of SOM relies on a competitive iterative process with a neighborhood function that acts as a smoothing kernel over the output layer \cite{SOM}. Typically, for each training sample, the best matching unit (BMU) is selected by using the Euclidean distance and then its local neighborhood is updated to make it slightly similar to the training sample. The neighborhood definition depends on the output layer. In our case, we use a regular quadrangular grid, but future improvements can be achieved by using more complex topologies such as toroidal or spherical grids \cite{Mount2011}. The hyperparameter of SOM is the number of output neurons. In the experimental section, neurons weights are initialized by using PCA.

\textbf{Growing Neural Gas (GNG)}: a common way to avoid the hyperparamer selection of SOM is to use growing structures that incrementally increase the number of neural units depending on the topology of the input data. GNG is an iterative algorithm to approximate the topology of a multidimensional dataset by using a changing number of neural units represented as a graph. In the most general form, the algorithm sequentially grows the nodes and adjusts the graph to the input data. In this way, each node of the graph has assigned a neural weight in the input space, and the algorithm sequentially adds or/and removes nodes based on cumulative error measurements between the nodes and the data \cite{Fritzke1995, Martinetz1991}. An important aspect of the GNG is the position of the first two nodes. In the experimental section, the first nodes are randomly located in the input space. {\color{black} Aditionally, the GNG maximum number of neurons is defined as $400$ and $900$ in seek of a fair comparison with $SOM_{20}$ and $SOM_{30}$, respectively.

For our particular interests, GNG and SOM play a similar role, and their usage in the global framework is the same; however, the predefined topology of SOM simplifies the understanding and visualization of the patterns captured by the algorithm in the application case. }

\subsection{Hyperparameters, classification rules, and post-learning evaluation} \label{sec:evaluationStrategies}

When evaluating manifold methods the most challenging part is to quantify if the patterns learned are modified by the phenomena under study.  Previous studies usually follow two different strategies: the first one quantifies the information lost when moving the training dataset from the original space to the intrinsic dimension \cite{Saxena2004}. The second strategy uses the manual labels or human knowledge to analyze the intrinsic dimension (output space) in a post-learning analysis \cite{Jing2011}.

In our case, the information strategy is used to define the hyperparameters of the Isomap and the SOM. In particular, we use the reconstruction error to select the number of neighbors of the Isomaps as proposed in \cite{Saxena2004}, and the Topological Conservation Quality (TCQ) to define the number of output neurons of SOM \cite{Arous2010a}.  In the particular case of SOM the TCQ is selected to include in the analysis the concept of temporal continuity preservation; However, a similar analysis can be obtained by using alternative evaluation criteria such as the topographic product \cite{Bauer1992}, or the topographic function \cite{Villmann1997}. In general, the TCQ measures the number of times that the SOM transformation breaks a contiguity in the input data. In the input space, we define as contiguous two consecutive frames. In the output space two neurons are contiguous if they share one border. Formally the TCQ is defined as (\ref{eq:topo}), where $Q$ is the number of training samples and $u(x_{q})=1$ if the two closest neurons of an input vector $x_q$ are contiguous in the output space, and $u(x_{q})=0$ otherwise.

\begin{eqnarray}\label{eq:topo}
    TCQ = \frac{\sum_{q=1}^{Q}u(x_{q})}{Q}
\end{eqnarray}

Once defined the hyperparameters, a post-learning analysis is done by using the manual labels to quantify the performance of the proposed manifold methods. For this purpose, each manifold method is trained on each global feature and dataset. Then a classification analysis is performed using the manual labels and defining as reference scores two popular supervised classifiers, namely Support Vector Machine (SVM) and Random Forest (RF). It is noteworthy that the supervised classifiers are in a favored position because they are theoretically developed to exploit the differences among manual labels; however, the closer the score of the manifold methods to the classifiers score, the more related the patterns learned are with the phenomena measured by the manual labels.

To use the manifold methods as classifiers, we use a majority voting rule in the output space (intrinsic dimension) using the training samples and their manual labels. For Isomaps and PCA, the majority voting rule is evaluated using the $10$ closest training frames in the output space. For SOM, the majority voting rule is evaluated on the training frames that activated the same output neuron of each testing sample.

\section{Experimental results}\label{sec:results}

This section evaluates the capabilities of the proposed manifold methods to capture light changes and separate different locations using global features.  In the first part of this section, we calibrate the hyperparameters of the Isomap and SOM while preserving the unsupervised nature of the training phase. Later, we use the manual labels to analyze the patterns learned under a classification approach \cite{Jing2011}. Finally, the discriminative ranking learned by a Random Forest is used to analyze the most relevant dimensions of the proposed global features. 

\subsection{Defining the hyperparameters}\label{sec:hyperparameters}

\begin{figure}[h!]
    \centering
    \includegraphics[width=0.8\linewidth]{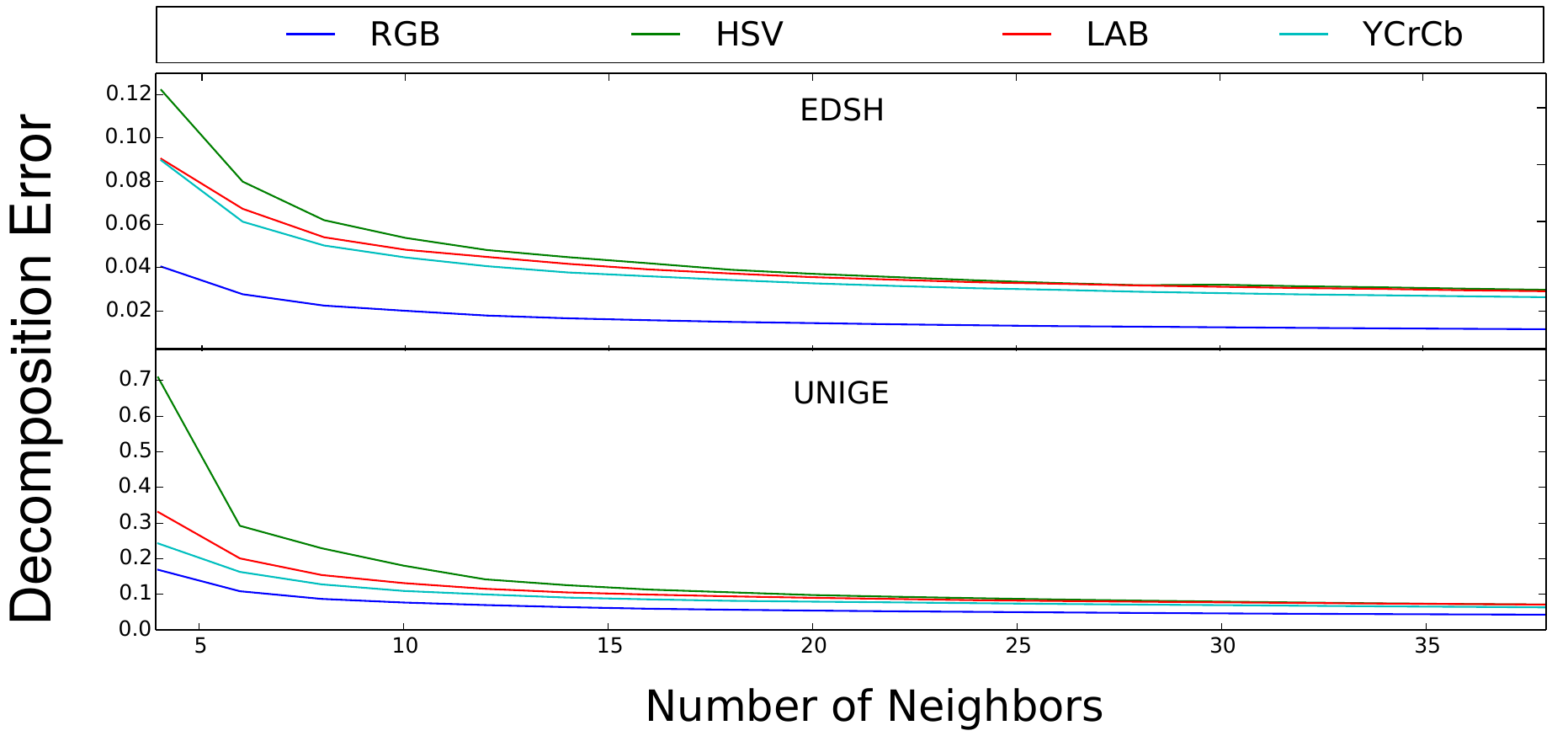}
    \caption{Isomap reconstruction error in function of the number of neighbors}
    \label{fig:isomap_kn}
\end{figure}

To define the number of neighbors considered in Isomaps we use the reconstruction error, which is the amount of information lost when transforming a point from the original space (global feature) to the intrinsic dimension. Figure \ref{fig:isomap_kn} shows the reconstruction error of the Isomap when the number of closest neighbors increases. Note that, for all the features; the reconstruction error starts stabilizing when the $12$ closest neighbors are used. Therefore, we use $1$2 as the parameter in the remaining part of the paper.

\begin{figure}[h!]
    \centering
    \includegraphics[width=0.8\linewidth]{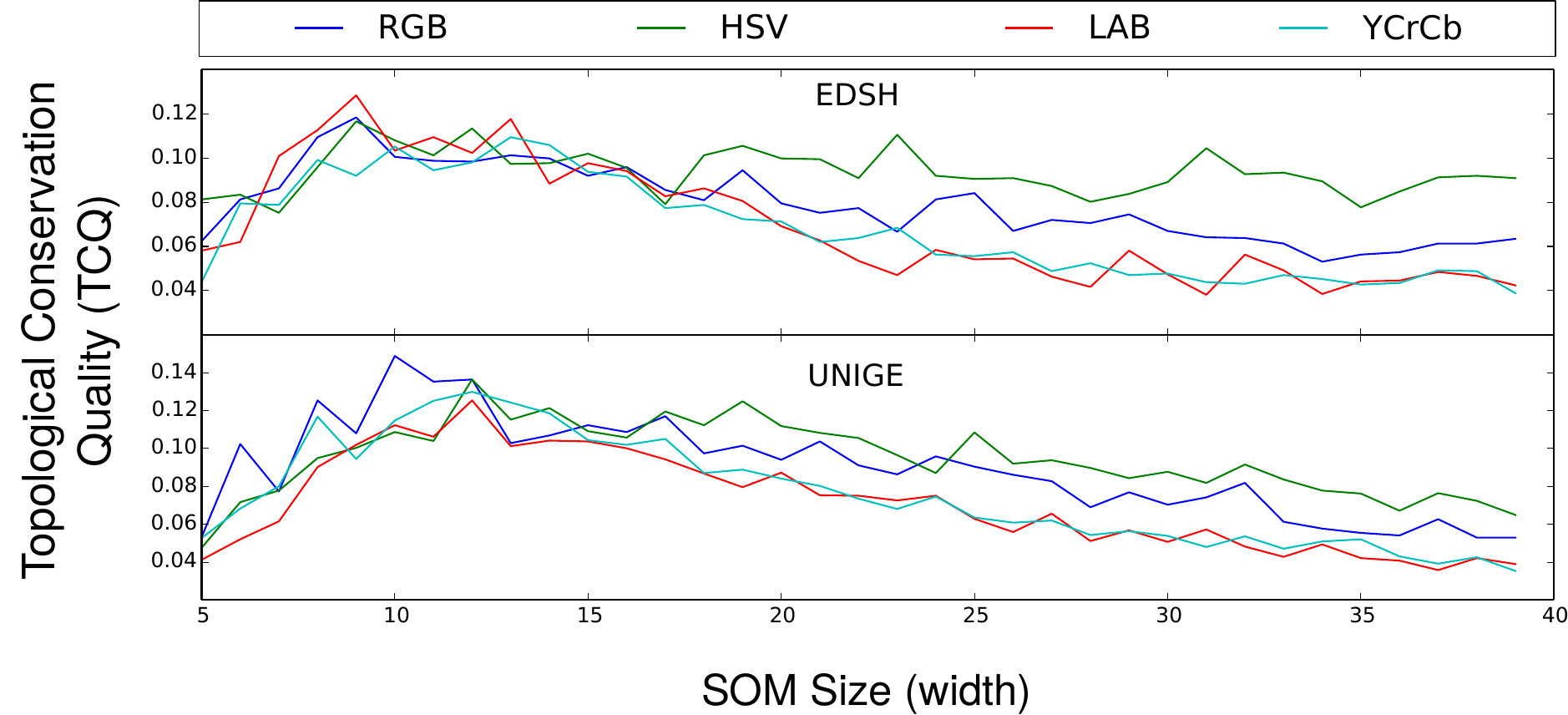}
    \caption{TCQ error in function of the number of neighbors}\label{fig:som_parameters}
\end{figure}

Regarding the number of output neurons of the SOM we use the TCQ, as defined in Section \ref{sec:evaluationStrategies}. Figure \ref{fig:som_parameters} shows the TCQ for different SOM sizes. Two findings are highlighted from the figure: i) A small number of neurons offers a topological advantage in the TCQ, because the fewer the output neurons to activate, the easier to preserve contiguities in the output space.  ii) The TCQ starts stabilizing for large SOMs, around $20\times20$ for EDSH and $30\times30$ for UNIGE dataset. In the experimental results we use three SOM sizes: $5\times5$, $20\times20$ and $30\times30$, denoted as $SOM_{5},SOM_{20},SOM_{30}$, respectively.

\subsection{Post-learning analysis}\label{sec:classification}

To evaluate the patterns found by the manifold methods we perform an exhaustive post-learning analysis under a classification framework using the manual labels and defining as reference scores the performance of SVM (linear kernel) and a RF (10 decision trees with maximum depth 10). For this purpose we define two different classification problems: i) Discriminate among indoors and outdoors frames ii) Classify the labeled locations given by the datasets (e.g. Kitchen, Office, Street, etc).

Table \ref{tab:supervisedEvaluation} shows the percentage of testing data successfully classified by each method (columns) when using different features (rows). The table contains two horizontal groups, one for each classification problem. The first group shows the performance for the binary problem (indoor/outdoor), and the second group shows the strict multiclass match for the detailed locations. The first group of columns shows the unsupervised methods while the second group shows the supervised classifiers results. Note that, despite not using manual labels in the training phase, the performance of the unsupervised methods are close to their supervised counterparts, which validates the patterns learned, and confirms the relationship between the proposed global features with the light/location conditions.

\begin{table*}
\centering
\scriptsize
\caption{\color{black} Supervised evaluation of different methods (columns) when used on top of different features (rows). Performance values are presented in two horizontal groups, one per classification problem (indoor/outdoor and location). The performance of both datasets (EDSH and UNIGE-Hands) are presented. The values reported are accuracy (properly classified frames over the total testing frames) for each feature/method combination.}
\begin{tabular}{>{\centering}m{0.1cm}>{\centering}m{0.3cm}>{\centering}m{0.8cm}|>{\centering}m{0.8cm}>{\centering}m{0.8cm}>{\centering}m{0.8cm}>{\centering}m{0.8cm}>{\centering}m{0.8cm}>{\centering}m{0.8cm}>{\centering}m{0.8cm}|>{\centering}m{0.8cm}r}
\toprule
      &       &       & \multicolumn{7}{c|}{Unsupervised}      & \multicolumn{2}{c}{Supervised} \\
\midrule
      & & feature & $SOM_{5}$ & $SOM_{20}$ & $SOM_{30}$ & $GNG_{400}$ & $GNG_{900}$ & PCA   & Isomap & SVM   & RF \\ \midrule
\multirow{10}[0]{*}{\begin{sideways}Indoor and Outdoor\end{sideways}} & \multirow{6}[0]{*}{\begin{sideways}EDSH\end{sideways}} 
                                                                         & RGB   & 0.679 & 0.799 & 0.781 & 0.765 & 0.757 & 0.745 & 0.742 & 0.790 & 0.849 \\
                                                                      &  & HSV   & 0.772 & 0.767 & \textbf{0.808} & 0.782 & \textbf{0.807} & 0.731 & 0.739 & 0.891 & 0.858 \\
                                                                      &  & LAB   & 0.686 & 0.773 & 0.730 & \textbf{0.804} & \textbf{0.813} & 0.656 & 0.773 & 0.843 & 0.839 \\
                                                                      &  & YCrCb & 0.614 & 0.616 & 0.610 & 0.729 & 0.718 & 0.626 & 0.619 & 0.763 & 0.782 \\
                                                                      &  & GIST  & 0.660 & \textbf{0.823} & 0.757 & \textbf{0.837} & \textbf{0.810} & 0.642 & 0.647 & 0.749 & 0.787 \\ \cline{2,-11}
               & \multirow{7}[0]{*}{\begin{sideways}UNIGE\end{sideways}} & RGB   & 0.902 & 0.925 & 0.923 & 0.872 & 0.885 & 0.637 & 0.666 & 0.923 & 0.971 \\
                                                                      &  & HSV   & 0.980 & \textbf{0.990} & \textbf{0.988} & 0.974 & \textbf{0.980} & 0.871 & 0.945 & 0.977 & 0.986 \\
                                                                      &  & LAB   & 0.912 & 0.961 & 0.947 & 0.977 & 0.947 & 0.778 & 0.957 & 0.979 & 0.988 \\
                                                                      &  & YCrCb & 0.775 & 0.891 & 0.894 & 0.942 & 0.969 & 0.772 & 0.931 & 0.971 & 0.975 \\
                                                                      &  & GIST  & 0.585 & 0.844 & 0.819 & 0.841 & 0.852 & 0.669 & 0.738 & 0.964 & 0.871 \\ \midrule
\multirow{10}[0]{*}{\begin{sideways}Location\end{sideways}}           & \multirow{6}[0]{*}{\begin{sideways}EDSH\end{sideways}} 
                                                                         & RGB   & 0.317 & 0.483 & 0.477 & 0.500 & 0.479 & 0.446 & 0.483 & 0.503 & 0.629 \\
                                                                      &  & HSV   & 0.417 & 0.493 & \textbf{0.558} & 0.534 & \textbf{0.557} & 0.319 & 0.339 & 0.551 & 0.669 \\
                                                                      &  & LAB   & 0.317 & 0.409 & 0.349 & 0.523 & 0.528 & 0.287 & 0.388 & 0.452 & 0.573 \\
                                                                      &  & YCrCb & 0.338 & 0.218 & 0.227 & 0.410 & 0.396 & 0.188 & 0.238 & 0.330 & 0.530 \\
                                                                      &  & GIST  & 0.423 & 0.532 & 0.519 & 0.535 & 0.518 & 0.286 & 0.348 & 0.554 & 0.517 \\ \cline{2,-11}
               & \multirow{7}[0]{*}{\begin{sideways}UNIGE\end{sideways}} & RGB   & 0.618 & 0.846 & 0.836 & 0.783 & 0.793 & 0.424 & 0.457 & 0.840 & 0.932 \\
                                                                      &  & HSV   & 0.826 & \textbf{0.954} & \textbf{0.963} & 0.934 & \textbf{0.957} & 0.672 & 0.831 & 0.954 & 0.954 \\
                                                                      &  & LAB   & 0.706 & 0.851 & 0.812 & 0.942 & 0.904 & 0.554 & 0.820 & 0.920 & 0.928 \\
                                                                      &  & YCrCb & 0.651 & 0.811 & 0.777 & 0.884 & 0.905 & 0.637 & 0.843 & 0.918 & 0.933 \\
                                                                      &  & GIST  & 0.307 & 0.688 & 0.644 & 0.661 & 0.719 & 0.354 & 0.404 & 0.881 & 0.708 \\
\bottomrule
\end{tabular}
\label{tab:supervisedEvaluation}
\end{table*}

In particular, Table \ref{tab:supervisedEvaluation} shows that within the unsupervised techniques the large SOM and GNG perform the best. The small differences between the SOM and GNG performance can be explained by the initialization of the neurons and the algorithmic differences. The first neurons of the GNG are located randomly in the input space while the SOM initial weights are defined by using PCA. The table also shows valuable insights about the most discriminative features. It is noteworthy the performance of the methods when HSV is used, particularly in the unsupervised approach. This fact confirms the intuition of previous works on which the use of HSV leads to algorithmic improvements when used as a proxy for the light conditions. About the datasets, it is possible to conclude that the EDSH dataset is the most challenging, especially for the location classification problem. Interestingly, in the Indoor/outdoor problem of EDSH dataset, the GIST achieves a good performance, but it is outperformed in the remaining problems by HSV.

\begin{figure}[h!]
    \centering
    \includegraphics[width=1\linewidth]{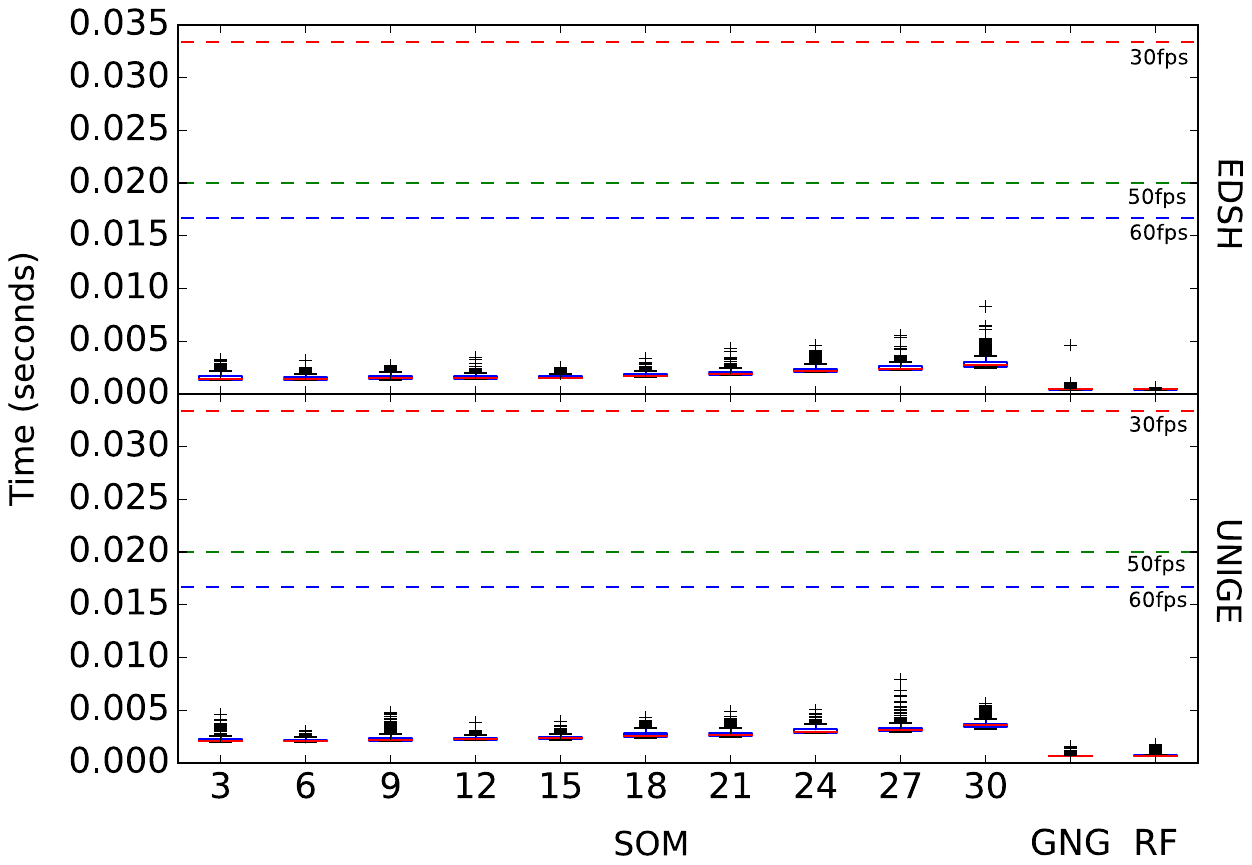}
    \caption{Execution time required different methods (Multiple SOM, GNG, RF) to transform a feature vector.}\label{fig:executionTime}
\end{figure}

More in detail Tables \ref{tab:confusionHSV_EDSH} and \ref{tab:confusionHSV_UNIGE} show the confusion matrix of the $SOM_{30}$ and the Random Forest for the EDSH and the UNIGE dataset, when HSV color space is used. As expected from Table \ref{tab:supervisedEvaluation} the location of the EDSH are more challenging, which creates larger confusion levels. {\color{black} This is the case, for example, of ``Stairs 1'' frames, which are frequently confused with kitchen frames by both algorithms due to the presence of a similar floor and wall color in both locations.}  Regarding the UNIGE dataset, a good performance is obtained in all the locations achieving values larger than $93\%$ for the unsupervised approach. The difference in the performances of both datasets shows the importance of having locations with enough data for a classification approach; however, it allows us to conclude the existence of structural similarities in the colour configuration and light conditions of the frames labelled as ``Kitchen'' and ``Stairs 1''. Figure \ref{fig:executionTime} shows the time required by different sizes of SOM, GNG and RF to transform a descriptor to the output space. The horizontal lines, from top to bottom, show the frequency required to achieve real-time performance on videos with 30, 50, and 60 frames per second respectively. There is a computational advantage in the speed of GNG and RF; however, all of them are fast enough to process $50fps$. The differences in performance can be a consequence of the particular implementations.

\begin{table*}
\scriptsize

\begin{subtable}{\textwidth}
\caption{Confusion matrix for the EDSH dataset location problem}
\centering
\begin{tabular}{rr|rrrrrrr|rrrrrrr}
\toprule
     &      & \multicolumn{7}{c|}{$SOM_{30}$}          & \multicolumn{7}{c}{RF} \\
\midrule
 & & \begin{sideways}floor1\end{sideways}  &  \begin{sideways}street\end{sideways}  &  \begin{sideways}lobby0\end{sideways}  &  \begin{sideways}kitchen\end{sideways}  &  \begin{sideways}stair2\end{sideways}  &  \begin{sideways}stair1\end{sideways}  &  \begin{sideways}lobby1\end{sideways}  &  \begin{sideways}floor1\end{sideways}  &  \begin{sideways}street\end{sideways}  &  \begin{sideways}lobby0\end{sideways}  &  \begin{sideways}kitchen\end{sideways}  &  \begin{sideways}stair2\end{sideways}  &  \begin{sideways}stair1\end{sideways}  &  \begin{sideways}lobby1\end{sideways} \\ \midrule
\multicolumn{1}{c}{\multirow{7}[0]{*}{\begin{sideways}HSV\end{sideways}}} &
                        floor1   & 0.511    &     0.173    &     0.170    &     0.061    &     0.015    &     0.033    &     0.036 & 0.675    &     0.046    &     0.191    &     0.064    &     0.012    &     0.003    &     0.009 \\
\multicolumn{1}{c}{} &  street   & 0.131    &     0.696    &     0.003    &     0.104    &     0.018    &     0.048    &     0.000 & 0.012    &     0.911    &     0.015    &     0.051    &     0.003    &     0.006    &     0.003 \\
\multicolumn{1}{c}{} &  lobby0   & 0.071    &     0.083    &     0.643    &     0.119    &     0.024    &     0.060    &     0.000 & 0.048    &     0.119    &     0.726    &     0.083    &     0.012    &     0.000    &     0.012 \\
\multicolumn{1}{c}{} &  kitchen  & 0.000    &     0.056    &     0.074    &     0.759    &     0.019    &     0.019    &     0.074 & 0.111    &     0.000    &     0.167    &     0.722    &     0.000    &     0.000    &     0.000 \\ 
\multicolumn{1}{c}{} &  stair2   & 0.000    &     0.000    &     0.078    &     0.206    &     0.451    &     0.010    &     0.255 & 0.020    &     0.000    &     0.039    &     0.324    &     0.392    &     0.000    &     0.225 \\
\multicolumn{1}{c}{} &  stair1   & 0.000    &     0.083    &     0.104    &     0.458    &     0.115    &     0.156    &     0.083 & 0.198    &     0.062    &     0.281    &     0.344    &     0.021    &     0.062    &     0.031 \\
\multicolumn{1}{c}{} &  lobby1  & 0.000    &     0.045    &     0.091    &     0.106    &     0.121    &     0.076    &     0.561 & 0.000    &     0.015    &     0.227    &     0.000    &     0.106    &     0.045    &     0.606 \\\bottomrule
\end{tabular}%
\label{tab:confusionHSV_EDSH}

\end{subtable}
\vspace{50px}
\begin{subtable}{\textwidth}
\caption{Confusion matrix for the UNIGE dataset location problem}
\centering
\begin{tabular}{rr|rrrrr|rrrrr}
\toprule
     &      & \multicolumn{5}{c|}{$SOM_{30}$}          & \multicolumn{5}{c}{RF} \\
\midrule
     &      & \begin{sideways}Bench\end{sideways}  &  \begin{sideways}Street\end{sideways}  &  \begin{sideways}Bar\end{sideways}  &  \begin{sideways}Office\end{sideways}  &  \begin{sideways}Kitchen\end{sideways}  &  \begin{sideways} Bench\end{sideways}  &  \begin{sideways}Street\end{sideways}  &  \begin{sideways}Bar\end{sideways}  &  \begin{sideways}Office\end{sideways}  &  \begin{sideways}Kitchen\end{sideways} \\ \midrule
\multicolumn{1}{c}{\multirow{5}[0]{*}{\begin{sideways}HSV\end{sideways}}} & 
                        Bench    & 0.931    &     0.036    &     0.014    &     0.004    &     0.014  &  0.859    &     0.123    &     0.011    &     0.000    &     0.007 \\ 
\multicolumn{1}{c}{} &  Street   & 0.025    &     0.946    &     0.021    &     0.008    &     0.000  &  0.008    &     0.967    &     0.013    &     0.008    &     0.004 \\ 
\multicolumn{1}{c}{} &  Bar      & 0.000    &     0.012    &     0.978    &     0.003    &     0.006  &  0.006    &     0.012    &     0.972    &     0.000    &     0.009 \\ 
\multicolumn{1}{c}{} &  Office   & 0.000    &     0.018    &     0.007    &     0.964    &     0.011  &  0.000    &     0.000    &     0.004    &     0.971    &     0.026 \\ 
\multicolumn{1}{c}{} &  Kitchen  & 0.000    &     0.010    &     0.000    &     0.000    &     0.990  &  0.000    &     0.003    &     0.000    &     0.000    &     0.997 \\ \bottomrule
\end{tabular}%
\label{tab:confusionHSV_UNIGE}
\end{subtable}
\caption{Confusion matrix of SOM and RF using the global HSV color space for the
location problem of the EDSH and UNIGE dataset.}
\end{table*}

Another intuitive way to analyze the results is by visualizing the learned patterns. In summary, a well performed dimensional mapping must locate frames close to each other, in the output space, if they are under similar light conditions and scene configuration. In other words, if the proposed features are related to the light/location conditions, the unsupervised method will try to separate them in the output space. The quality of that separation is ruled by the complexity of the data and the manifold method used.

\begin{figure*}
    \centering
    \begin{subfigure}[t]{0.9\textwidth}
        \includegraphics[width=0.9\linewidth]{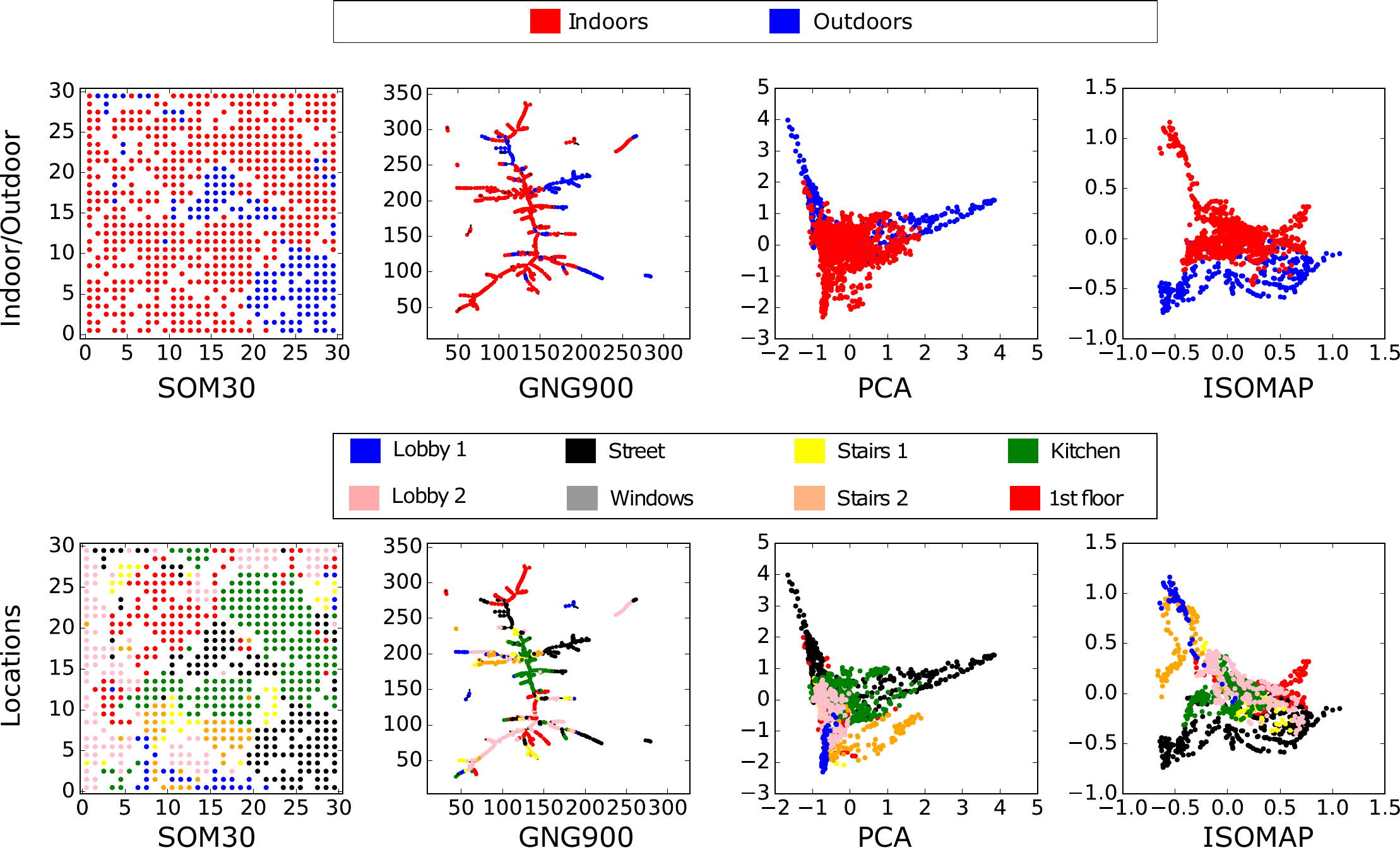}
	    \caption{Manifold output for EDSH dataset}
	\end{subfigure}\\
    \begin{subfigure}[t]{0.9\textwidth}
        \includegraphics[width=0.9\linewidth]{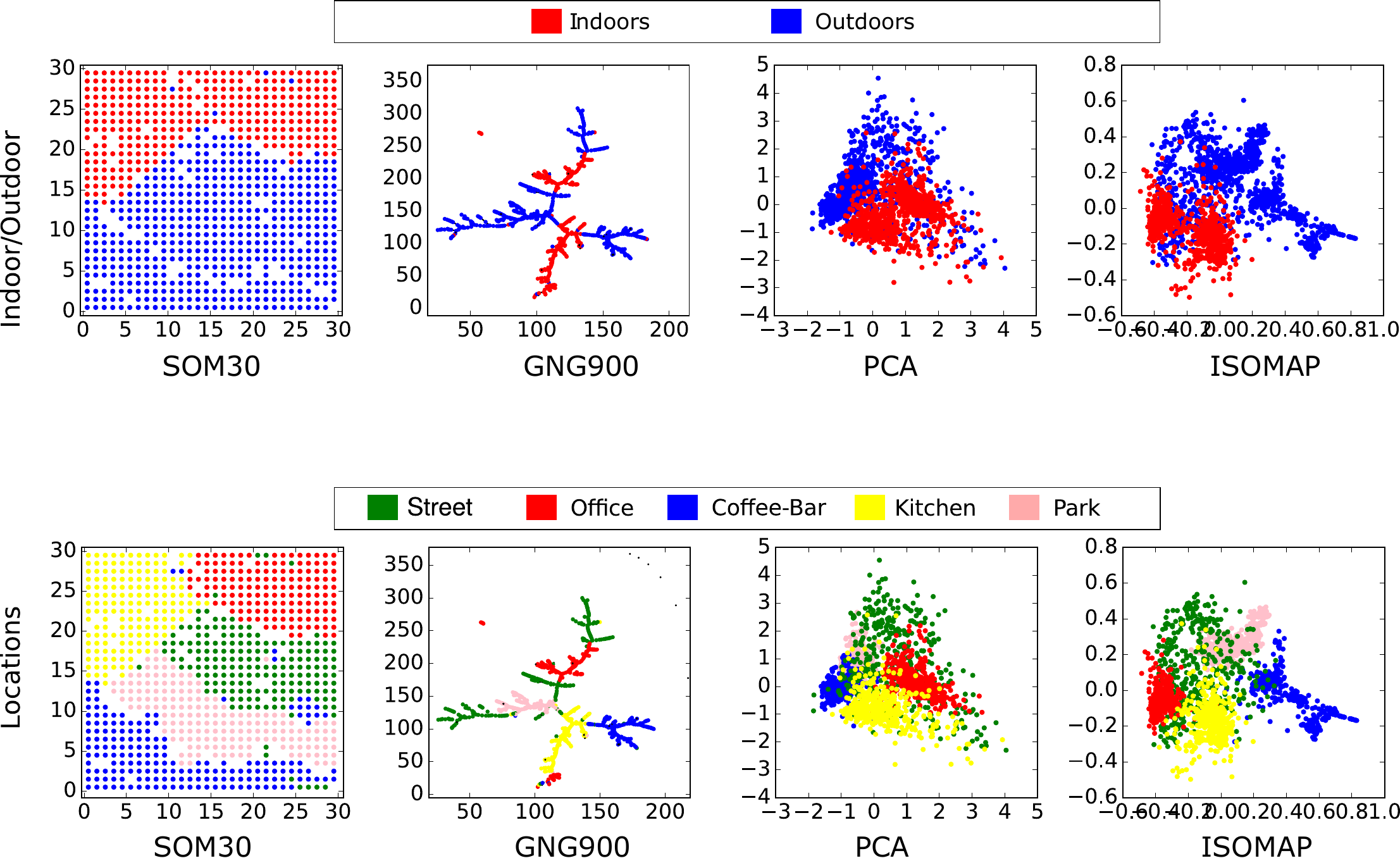}
	    \caption{Manifold output for UNIGE dataset}
        \label{fig:2dmapunige} 
    \end{subfigure}%
    \caption{\color{black}2D representation of the datasets using SOM, GNG, PCA, and Isomaps, for the EDSH (a) and UNIGE (b) datasets.}

    \label{fig:SOM2}
\end{figure*}

Figure \ref{fig:SOM2} shows the 2D output for the $SOM_{30}$, $GNG$, $Isomap$, and $PCA$, for both datasets using HSV. Different colors represent the manual labels. In the case of SOM and GNG, each neuron is labeled with the majority voting of the neural activations hits. The figure clearly shows that SOM successfully groups similar inputs in the same regions of the output layer. The GNG also create some groups of neurons for each location, but its visualization makes difficult to conclude. In the case of PCA and Isomaps, the patterns in the output space are not so evident, but definitely, the non-linearity of Isomaps allows them to capture more information than PCA, which is clearly affected by the orthogonality of the intrinsic dimensions.

It is remarkable the output space of the $SOM_{30}$ in the UNIGE dataset, on which both classification problems are located in different parts of the output layer. For the EDSH dataset, it is also possible to delineate some clusters, such as the kitchen (green) the street (black), the $1st$ floor (red) and the stairs (yellow and orange). However, the remaining locations are not easily visible, e.g., both lobbies (in blue and pink). This is explained by the small number of frames available for these locations in the dataset.

\begin{figure*}
\centering
    \includegraphics[width=1\linewidth]{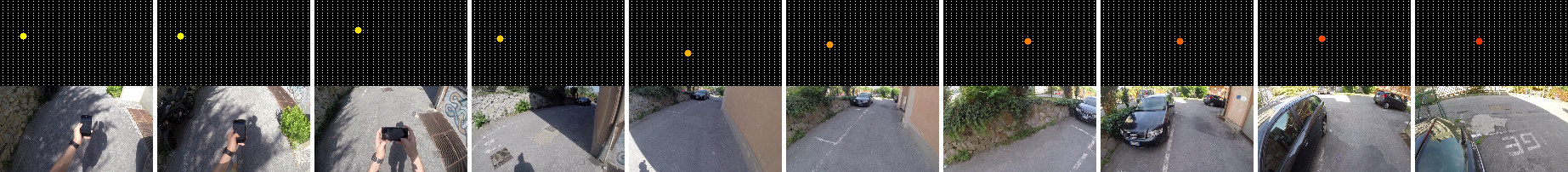}
    \caption{$SOM_{30}$ signature for 40 seconds from the street video in the UNIGE
    dataset.  The first row shows the activated neurons in the SOM output layer by
    the frame presented in the lower row.}
\label{fig:signature}
\end{figure*}

Figure \ref{fig:signature} shows the $SOM_{30}$ signature when transforming a uniform sampling of $40$ seconds from the street video of the UNIGE dataset using HSV. In the first row are the activated neurons (unsupervised locations) ordered by time from left to right. In the second row are the compressed snapshots for the input frames. As can be seen from the first row, the $SOM_{30}$ activations start on the left side and moves to the middle of the grid while the user walks in the street through different light conditions. The point color represents the temporal dimension, being yellow the first frame and red the last one.

\subsection{Feature Analysis}

This subsection exhaustively analyzes the discriminative capabilities of the proposed global features and combine the most relevant dimensions to improve the dimensional mapping. For this purpose we follow two steps: i) The global features (RGB, HSV, LAB, YCrCb, GIST) are combined and used to train a RF on each dataset and classification problem described in Section \ref{sec:results}.  ii) The discriminative importance learned by the RF is exploited by adding, in order of importance, each of the original dimensions while evaluating the performance of RF and $SOM_{30}$.

\begin{figure}[h!]
    \centering
    \includegraphics[width=1\linewidth]{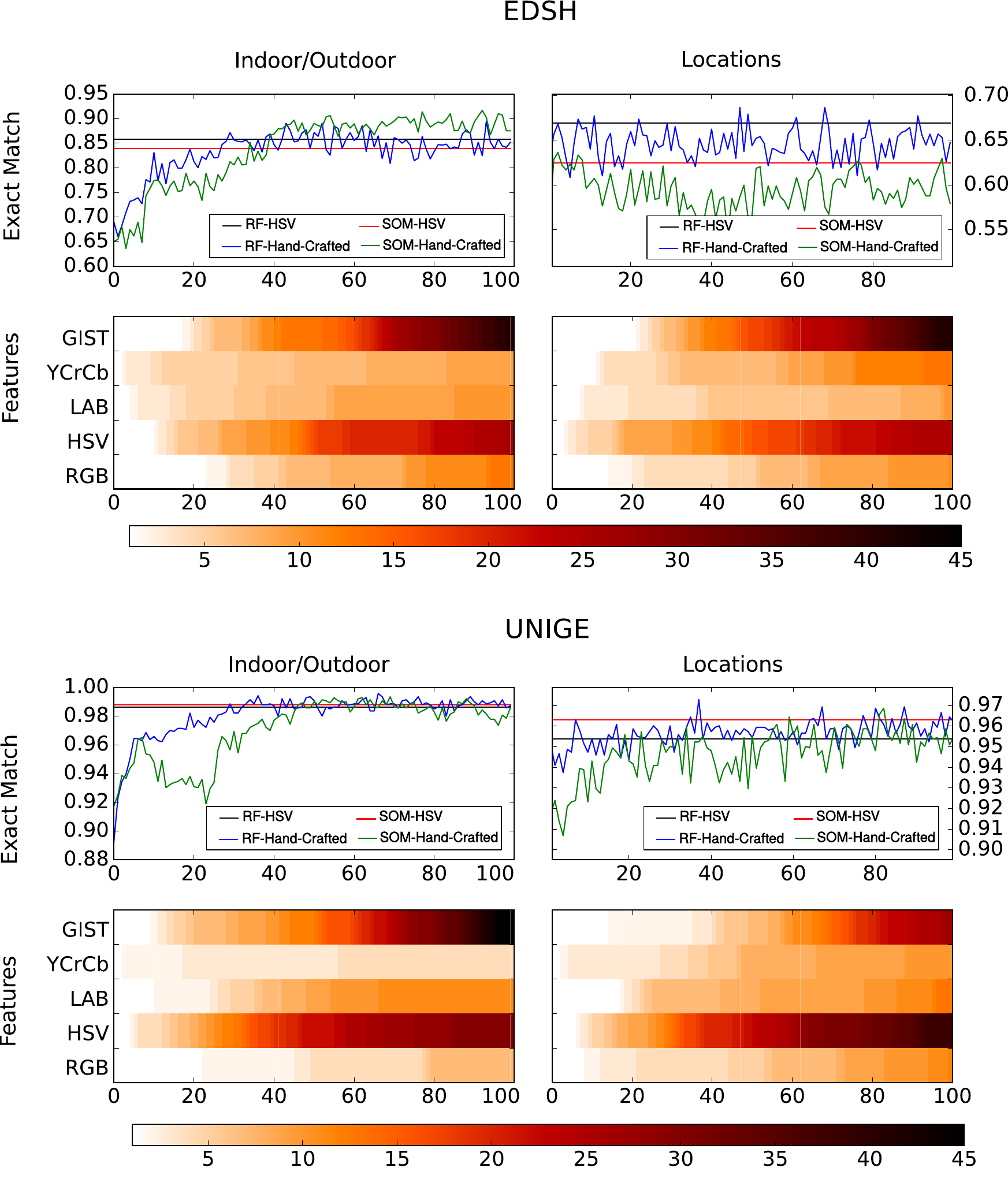}
    \caption{Performance of combined feature by adding on component by
    step: \textbf{top:}  EDSH dataset \textbf{bottom: } UNIGE dataset. First
    column shows the indoor/outdoor problem and second column visualizes the location
    problem (Section \ref{sec:results}). The lines plot represents performance
    and the heat maps represent the number of components selected in each step
    from each original feature. The color bars below the heat maps show the
    legend relating a color with a particular number. }
    \label{fig:manualFeature}
\end{figure}

Figure \ref{fig:manualFeature} summarizes the changes in performance (line plot) and the number of components (heat-map) belonging to each global feature on each step  (x-axis). The upper and  lower parts of the figure show the results for the EDSH and the UNIGE dataset, respectively. The first column corresponds to the indoor/outdoor problem and the second column to the location problem. The constant values in the line plots are the performance of $SOM_{30}-HSV$ and $RF-HSV$ reported in Table \ref{tab:supervisedEvaluation}.

From Figure \ref{fig:manualFeature} it is possible to conclude that combined features could improve the performance in the proposed classification problems.  For instance, for the EDSH dataset, the combined features improves the SOM accuracy from $84.7\%$ to $91.4\%$ and $62.1\%$ to $65.2\%$ in the indoor/outdoor and location problem, respectively.  For the UNIGE dataset, due to the original performance, the improvement is not as significant. However, for some steps in the location problem, the combined features reaches an accuracy of $99.2\%$, which is slightly better than the $98.7\%$ of the HSV version. It is also noteworthy the result on the location problem for the EDSH dataset, on which the combined feature is close to the SOM-HSV combination, but is not able to improve its performance considerably.  The latter fact confirms that the location problem in the EDSH dataset is the most challenging, not only for the manifold methods but also for the supervised classifiers.

{\color{black} Regarding the composition of the combined features, it is notable that by using less than $40$ components, it is possible to achieve similar performance to the SOM-HSV, which originally uses $94$ components. Additionally, for all cases, the method starts using HSV, YCbCr and LAB components as the most discriminative, but around the $30$ to the $40$ step, it aggressively uses GIST components to disambiguate the most difficult cases. It is important to note that HSV, YCbCr, and Lab, are color spaces designed to use one of the components for lumma and the other two components for chromatic information. A quick analysis of the GIST components suggests that the RF searches for orientations and scale in the scene. Finally, the RGB color-space is barely used.}

\section{Application case: Multi-model hand-detection}\label{sec:application}

Once confirmed the capabilities of SOM to capture light conditions and the global characteristics of the scene, its output can be used as a map of unsupervised locations to build a multi-model approach to different problems such as object recognition, hand-detection, video-summarization, activity recognition, among others. This section illustrates the use of the unsupervised layer by using the hand-detection problem as defined in \cite{Betancourt2015}, on which a Support Vector Machine (SVM) is trained with Histogram of Oriented Gradients (HOG) to detect whether the hands are being recorded by the camera or not \cite{Betancourt2015, Betancourt2014a}. The following part of this section uses the UNIGE dataset due to the intentional composition of frames with and without hands.

The hand-detection problem is used as example due to two reasons: i) It solves a simple question which makes it possible to illustrate the role of the unsupervised layer in the reported improvements; ii) The manual labeling is simple and easy to replicate. The proposed application can be extended to other hierarchical levels such as hand-segmentation; however, it would require extra labeling to supply quadratic growth of the number of neurons.

Our approach extends the method proposed in \cite{Betancourt2015} by training one hand-detector for each unsupervised neuron of the HSV-SOM described in Section \ref{sec:method}. Let's denote each neuron $i \in SOM_N$ and its local hand-detector as $hd_i^N$, and the global hand-detector as $hd^N$. Given an arbitrary frame $f$, the local and global confidence about the hand presence is given by the SVM probabilistic notation as stated in equation (\ref{eq:localModel}) and (\ref{eq:globalModel}), respectively. The model with the higher confidence is used to take the final decision. Here $\Theta$ refers to the hyperplane learned by the HOG-SVM when trained on the whole training dataset, and $\theta_i$ to the hyperplane obtained with a HOG-SVM when trained on local training set assigned to neuron $i$, which contains the training frames for which neuron $i$ was the best matching unit. Additionally, for each neuron $i$ a local testing set (LTS) is defined by combining the activations of the neighbouring neurons. The LTS of each neuron is used to evaluate its local $F1$-score. Due to the finite number of training frames, some neurons does not reach enough training frames or get only positive/negative frames which makes impossible to train their local hand-detectors. These neurons and the ones with local $F1$-score lower than $0.75$ are defined as degraded, and their hand-detector is replaced by the global version.

\begin{eqnarray}
    hd^N_i(f) = SVM(HOG(f)|\theta_i) \label{eq:localModel} \\
    hd^N(f) = SVM(HOG(f)|\Theta) \label{eq:globalModel}
\end{eqnarray}

\begin{figure}[h!]
\centering
    \includegraphics[width=1\linewidth]{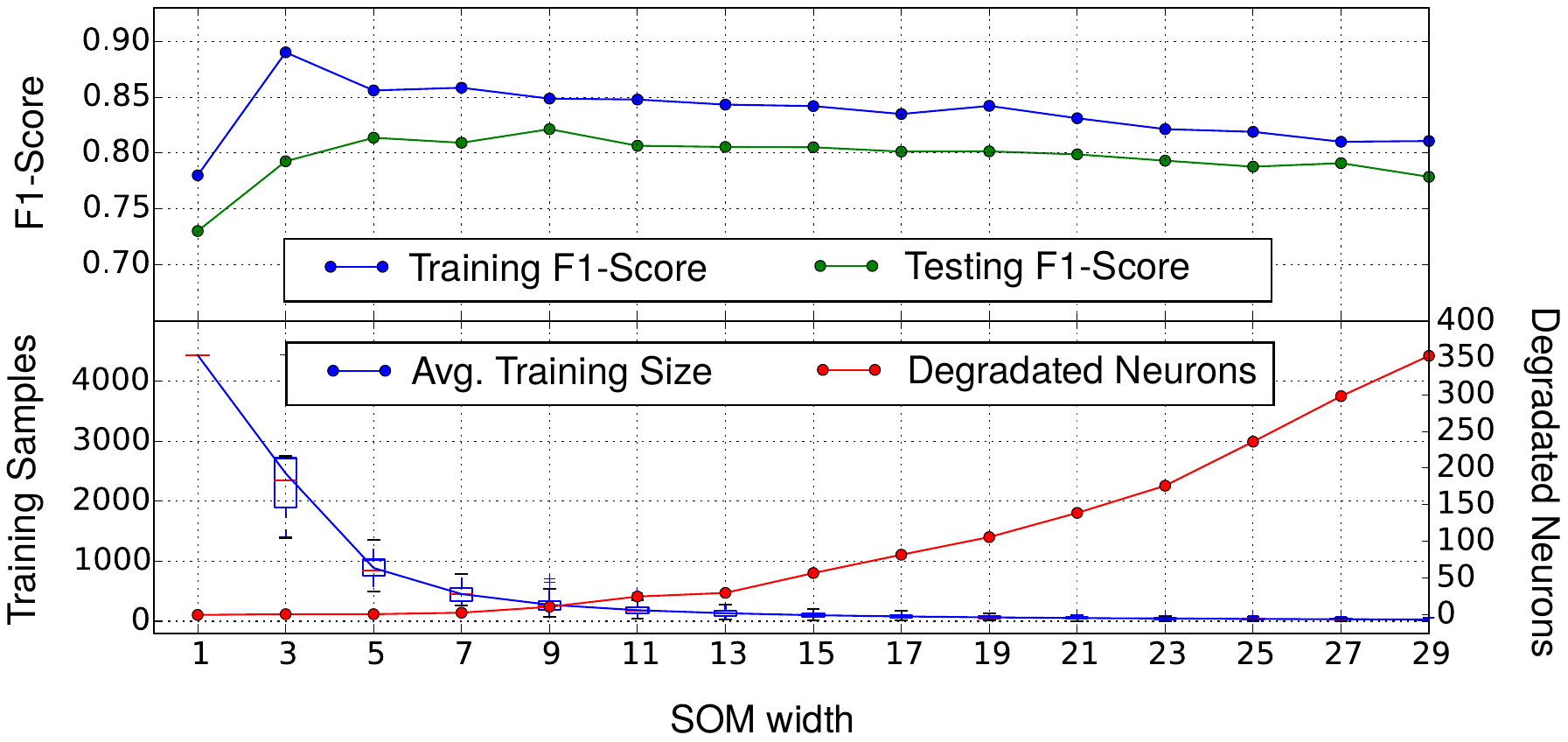}
    
    \caption{The upper part of the figure shows the training (blue) and testing
    (green) $F1$ score. The lower part shows the average number of training
    frames (blue) used to train $hd_i^N \in SOM_N$, and the number of
    degradated neurons (red). The horizontal axis is the size of the SOM.}

    \label{fig:sizeVsF1}
\end{figure}

Figure \ref{fig:sizeVsF1} summarizes the performance of the multi-model approach for different SOM sizes (x-axis). The upper half of the figure shows the training and testing $F1$ scores. This figure shows a quick increase in the $F1$ score which stabilizes for SOMs with more than $9^2$ neurons. The lower half of the figure shows the average number of training frames per neuron (blue) and the number of degraded neurons (red). Two important conclusions can be drawn from these figures: i) The multi-model approach overfits the training dataset on large SOMs ii) The number of degraded neurons increases quickly and, as a consequence, no extra benefit is obtained from larger SOMs. 

\begin{figure}[h!]
	\centering
	\begin{subfigure}[t]{0.24\textwidth}
		\includegraphics[width=\textwidth]{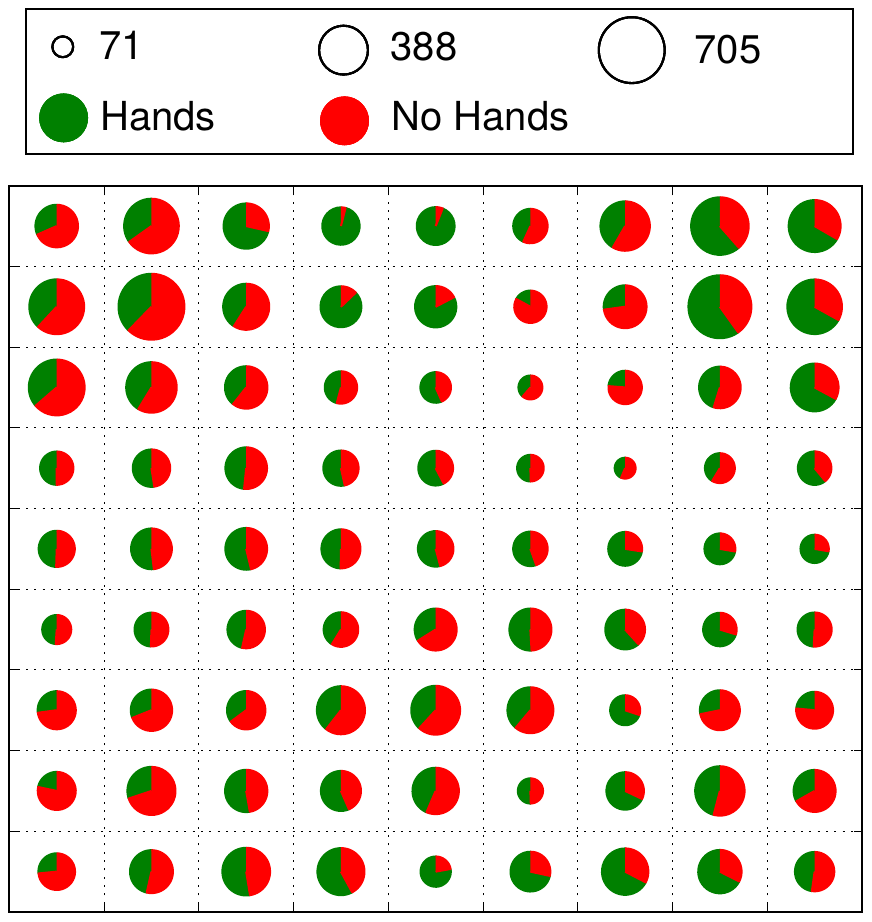}
		\caption{Frames with and without hands in the local training subset of each neuron. The size of each plot is given by the size of its local training subset.}
		\label{fig:trainingProportions}
	\end{subfigure}%
	~ 
	\begin{subfigure}[t]{0.24\textwidth}
		\includegraphics[width=\textwidth]{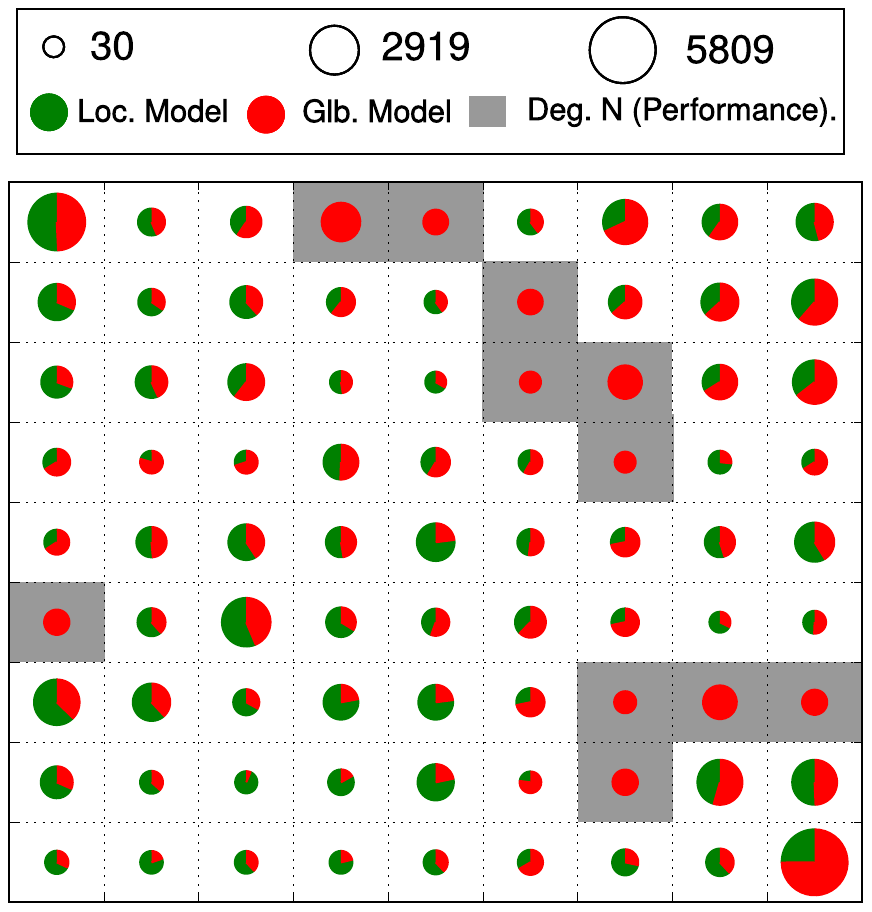}
		\caption{Localized model vs global model usage. The size each plot is the number of activations of the testing frames for each neuron.}
		\label{fig:modelUsage}
	\end{subfigure}%

	\begin{subfigure}[t]{0.24\textwidth}
		\includegraphics[width=\textwidth]{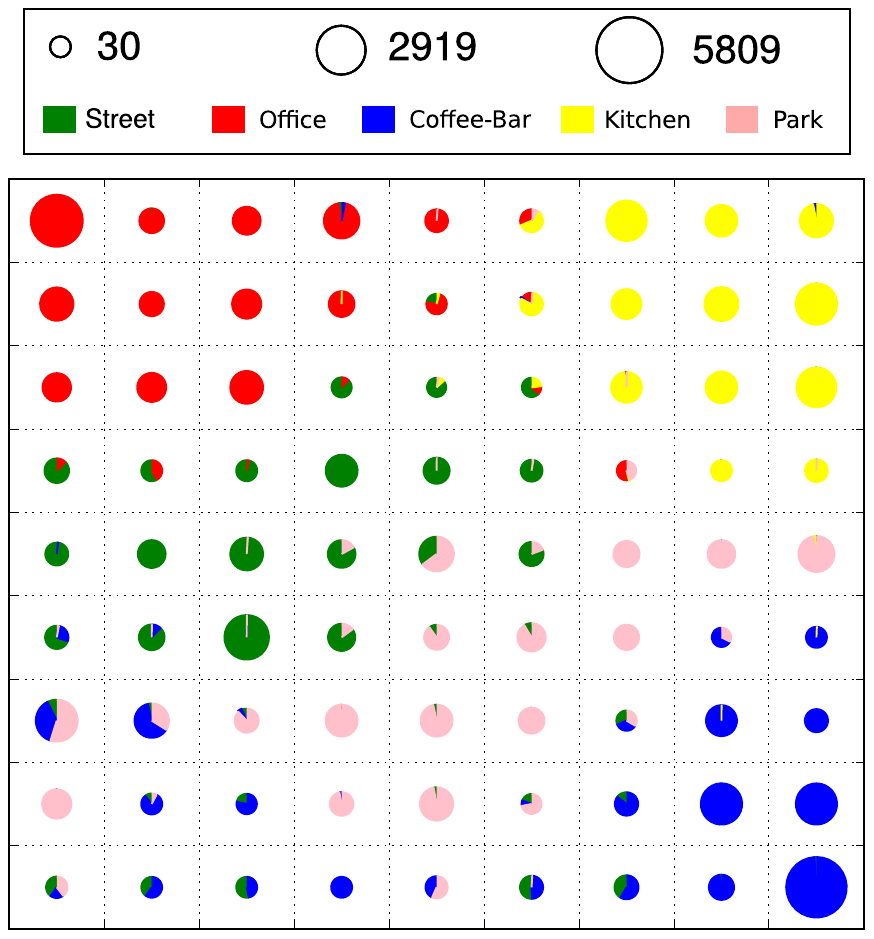}
		\caption{Composition of the testing frames for each neuron. The size of each plot is the number of activations of the testing frames for each neuron.}
		\label{fig:compositionUsage}
	\end{subfigure}%
	~ 
	\begin{subfigure}[t]{0.24\textwidth}
		\includegraphics[width=\textwidth]{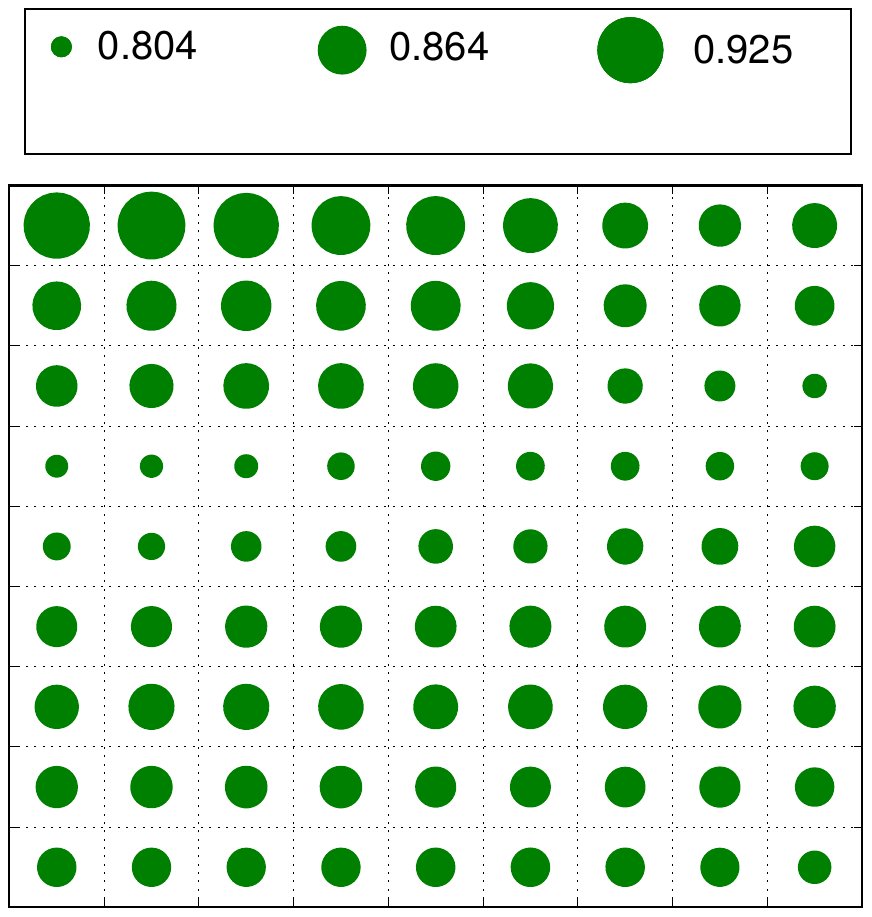}
		\caption{Testing performance by neuron (F1-score).}
		\label{fig:neuronPerformance}
	\end{subfigure}%
	\caption{Important facts about the mutimodel approach when using $SOM_9$ as model based.}
	\label{fig:proportions}
\end{figure}


\begin{table}[h!]
    \setlength{\tabcolsep}{4pt} 
\scriptsize
\centering
    \caption{True-Positives and True-Negatives comparison between a unique model approach as proposed in  \cite{Betancourt2015} (HOG-SVM) and a Multimodel approach using the $SOM_9$ (Ours)} \label{tab:onevsmulti}
    \begin{tabular}{lcc|cc"cc}
    \toprule
     & \multicolumn{ 2}{c|}{ True positive rate } & \multicolumn{ 2}{c"}{True negatives rate} & \multicolumn{ 2}{c}{\textbf{F1-score}}\\ \midrule
     & \multicolumn{1}{c}{HOG-SVM} & \multicolumn{1}{c|}{Ours} & \multicolumn{1}{c}{HOG-SVM} & \multicolumn{1}{c"}{Ours} & \multicolumn{1}{c}{\textbf{HOG-SVM}} & \multicolumn{1}{c}{\textbf{Ours}}\\ \midrule
    Office          & 0.888   &   \textbf{0.914}   &   0.928   &   \textbf{0.937} & 0.897 &  \textbf{0.917} \\  
    Street          & 0.767   &   \textbf{0.797}   &   0.871   &   \textbf{0.927} & 0.814 &  \textbf{0.856} \\  
    Bench           & 0.743   &   \textbf{0.799}   &   0.964   &   \textbf{0.966} & 0.832 &  \textbf{0.868} \\  
    Kitchen         & 0.618   &   \textbf{0.646}   &   0.773   &   \textbf{0.794} & 0.691 &  \textbf{0.718} \\  
    Coffee bar      & 0.730   &   \textbf{0.805}   &   0.695   &   \textbf{0.767} & 0.718 &  \textbf{0.790} \\  \midrule
    Total           & 0.739   &   \textbf{0.783}   &   0.846   &   \textbf{0.877} & 0.780 &  \textbf{0.821} \\  \bottomrule
    \end{tabular}
\end{table}

Table \ref{tab:onevsmulti} compares the performance of the HOG-SVM and the multi-model strategy on a $SOM_9$. The table shows the true-positive rate, true-negative rate and the $F1$ score for each location in the dataset.  In general, our approach considerably improves the performance for all locations, totalizing an improvement of $4.1$ $F1$ score points in the whole dataset. The location with the larger improvement is the \emph{Coffee-bar} with an increase of $7$ points in the $F1$ score. This improvement is explained by an increase of $7.5$ and $7.2$ percentual units in the true-positive and true-negative rate, respectively.

Finally, Figure \ref{fig:proportions} summarizes some neural characteristics of the $SOM_9$: Figure \ref{fig:trainingProportions} shows the number of training frames used per neuron and the proportions between frames with (green) and without (red) hands. Some neurons have a slightly unbalanced training. This fact is included in the hand-detector training phase by using these proportions as the weights of the class in the SVM. Figure \ref{fig:modelUsage} summarizes the use of $hd_i^9$ and $hd^9$. The size of the circle represents the number of testing frames activating a particular neuron. In turn, each circle is proportionally divided in green and red according to the number of times that the local or global model is used, respectively. The gray cells are the degraded neurons on which only the global model is always used. For this particular SOM size the degraded neurons are consequence of poor local $F1$ scores. Figure \ref{fig:compositionUsage} shows the composition of testing frames on each neuron in terms of its location. Note that, the resulting regions are in line with the regions presented in Section \ref{sec:classification}, Figure \ref{fig:2dmapunige}. Finally, Figure \ref{fig:neuronPerformance} shows the testing $F1$ score of each neuron.  It is noteworthy that the smallest $F1$ scores are located in a contiguous region of the $SOM_9$. This fact can be exploited by using a windowing to fuse the local models. In sake of an easy explanation of the application case, this improvement is not included in the current implementation.

\section{Conclusions and future research}\label{sec:conclusions}

This paper proposes an unsupervised strategy to endow wearable cameras with contextual information about the light conditions and location recorded by using global features. The main finding of our approach is that using SOM and HSV, it is possible to develop an unsupervised layer that understands the illumination and location characteristics on which the user is involved. Our experiments validate the intuitive findings of previous works using HSV global histograms as a proxy for the light conditions recorded by a wearable camera. As an application case, the unsupervised layer is used to face the hand-detection problem under a multi-model approach. The experiments presented in the hand-detection application considerably outperform the method proposed in \cite{Betancourt2015}.

The experimental results analyze the capabilities of different unsupervised methods to capture light and location changes in egocentric videos. The experimental results show that SOM can extract valuable contextual information about the illumination and location from egocentric videos without using manually labeled data.  

Regarding the relationship between the global features and the recorded characteristics, our experiment points at HSV as the color space having the most discriminative power. Additionally, it is shown that by following a simple feature selection, it is possible to obtain a combined feature, mainly formed by HSV and GIST, which makes easier for SOM to capture these patterns. Two issues about the combined feature to be accounted for: i) it is computationally expensive compared with using just HSV; ii) it indirectly introduces a dependence between the manual labels and the training phase.

Concerning future work, several challenges in the proposed method can be faced. One of the more promising is the use of deep features to extract more complex contextual patterns. This type of approach could considerably improve the scalability of the system in particular when the user is visiting multiple and unknown locations. This strategy could be considered an example of knowledge transfer on which the information about scene recognition is obtained from the neural coefficients obtained with non-wearable camera. Important considerations mentioned before must be accounted if deep features are included. In the application case, important improvements can be achieved if the proposed framework is applied to other hierarchical levels, for example, the unsupervised layer can be used to switch between different color spaces at a hand-segmentation level or used to select different dynamic models at a hand-tracking level \cite{Betancourt2016b}. Another interesting improvement to the current approach is to include dynamic information in the activated neurons by exploiting the temporal correlation and avoiding to execute the unsupervised method for each frame in the video stream \cite{Betancourt2015}. 

Finally, an interesting application of the proposed approach can be found in video summarization, visualization and captioning. In this line, the output space can be used to find easily and retrieve video segments recorded on similar locations or light conditions.


\section{Acknowledgment}

This work was partially supported by the Erasmus Mundus joint Doctorate in
Interactive and Cognitive Environments, which is funded by the EACEA, Agency of
the European Commission under EMJD ICE. Likewise, we thank the AAPELE
(Architectures, Algorithms and Platforms for Enhanced Living Environments) EU
COST action IC1303 for the STSM Grant, the International Neuroinformatics
Coordinating Facility (INCF) and the Finnish Foundation for Technology
Promotion (TES).

The authors thank the Cyberinfrastructure Service for High Performance
Computing, ``Apolo'', at EAFIT University, for allowing us to run our
computational experiments in their computing centre. 



\section*{Bibliography}
\footnotesize
\bibliographystyle{IEEEbib}
\bibliography{biblio,biblio2}

\end{document}